# A systematic review of trial-matching pipelines using large language models

Braxton A. Morrison[1#], Madhumita Sushil[2], Jacob S. Young[2#]

[1]Department of Medicine, University of California San Francisco, San Francisco, CA, USA
[2]Department of Neurological Surgery, University of California San Francisco, San Francisco, CA, USA

[#]Authors for correspondence
Jacob S. Young, MD
Assistant Professor
Principal Investigator, Brain Tumor Center
Department of Neurological Surgery
University of California, San Francisco
jacob.young@ucsf.edu

Braxton A. Morrison, MS
School of Medicine
University of California, San Francisco
braxton.morrison@ucsf.edu

## Abstract
Matching patients to clinical trial options is critical for identifying novel treatments, especially in oncology. However, manual matching is labor-intensive and error-prone, leading to recruitment delays. Pipelines incorporating large language models (LLMs) offer a promising solution. We conducted a systematic review of studies published between 2020 and 2025 from three academic databases and one preprint server, identifying LLM-based approaches to clinical trial matching. Of 126 unique articles, 31 met inclusion criteria. Reviewed studies focused on matching patient-to-criterion only (n=4), patient-to-trial only (n=10), trial-to-patient only (n=2), binary eligibility classification only (n=1) or combined tasks (n=14). Sixteen used synthetic data; fourteen used real patient data; one used both. Variability in datasets and evaluation metrics limited cross-study comparability. In studies with direct comparisons, the GPT-4 model consistently outperformed other models—even finely-tuned ones—in matching and eligibility extraction, albeit at higher cost. Promising strategies included zero-shot prompting with proprietary LLMs like the GPT-4o model, advanced retrieval methods, and fine-tuning smaller, open-source models for data privacy when incorporation of large models into hospital infrastructure is infeasible. Key challenges include accessing sufficiently large real-world data sets, and deployment-associated challenges such as reducing cost, mitigating risk of hallucinations, data leakage, and bias. This review synthesizes progress in applying LLMs to clinical trial matching, highlighting promising directions and key limitations. Standardized metrics, more realistic test sets, and attention to cost-efficiency and fairness will be critical for broader deployment.

## Acknowledgements
Original art is the work of Mr. Kenneth Probst, Department of Neurosurgery, UCSF. The authors used the Elicit application (Ought, https://elicit.org) during the early scoping stage to roughly identify which articles contained certain types of information and to help organize thematic areas for review. UCSF Versa was used to assist with conciseness. No text or analysis generated by the tool appears in the submitted manuscript; all synthesis, interpretation, and writing were performed by the authors.

## Main Text

### *Introduction*

*Clinical Trial Enrollment and Matching*

Clinical trials are essential for identifying novel treatment options and offering patients access to potentially life-saving therapies. This is particularly important in fields like oncology, where alternative therapeutic options may be limited.[1–3] One key stage of trial-matching includes patient recruitment, which represents ~32% of clinical trial cost and is the reason most frequently cited for the discontinuation of randomized controlled trials.[4,5] Another important stage is manual trial matching, which requires intensive review of complex eligibility criteria across extensive patient records to assess patient suitability for a given trial, a time-consuming and labor-intensive process.[6–8] This stage is also costly; one study reported a cost between $129.15 to $336.48 per enrolled patient.[8] Equity is another key issue with trial enrollment; minorities, elderly people and rural groups are often underrepresented in cancer trials, a problem which could potentially be addressed by screening all patients for a large number of trials.[9] Thus, screening more patients for eligibility across more trials could improve patient recruitment and equity, but cost, labor and time are core limiting factors to its implementation.

*Automated Clinical Trial Matching*

Automated matching systems aim to address these inefficiencies, with early approaches relying on rigid, rule-based methods. One approach involved generating queries from trial eligibility criteria that could be applied to identify potentially eligible patients in clinical databases.[10,11] Another involved extracting key details from patient records and converting them to a structured format to filter lists of trials.[12] Earlier rule-based automation systems—such as the Watson for Clinical Trial Matching—demonstrated potential, as seen in a 2016 Mayo Clinic pilot showing an 80% increase in enrollment for breast cancer trials.[13] These systems often lacked generalizability across institutions, patient populations, or trial protocols. Recently, large language models (LLMs) have emerged as a flexible alternative capable of interpreting both structured and unstructured data, extracting eligibility criteria, and matching patients to trials with greater scalability.

Synthesizing recent work from 2020 to 2025, we characterize datasets, common pipeline structures, comparative model performance, and evaluation strategies. We aim to provide a practical roadmap for researchers and implementers to understand the capabilities, limitations, and emerging best practices for deploying LLMs in this domain.

### *Results*

*Model Categories*

This review covered four main types of trial matching – “patient-criterion matching”, in which a patient is assessed for one or more criteria individually; “patient-to-trial,” through which a list of trials is determined for a given patient; “trial-to-patient,” through which a list of potentially eligible patients is provided for a given trial; and “binary classification”, in which it is determined whether a patient-trial pair is a match. Of the reviewed articles, four focused on matching patient-to-criterion only, ten on patient-to-trial only, two on trial-to-patient only, one on binary eligibility classification only and fourteen on combined tasks (**Table 1**). While in theory, a perfect match between a patient and all trial criteria should imply eligibility, real-world applications are more complex. Strict 100% patient-criterion match thresholds often exclude many patients who may still be eligible under fewer, more practical constraints. In a study by Gupta et al., they found that weighting more important criteria more heavily resulted in better performance in translating criterion-level assessments into patient-trial matches.[14]

*Model Pipeline Design*

Most LLM-based trial-matching algorithms involve four stages: (1) acquisition of patient and trial data; (2) data pre-processing; (3) retrieval of relevant information from patient records; and (4) matching patients to trials (**Fig. 2A**).

*Data Sets*

Pipeline inputs varied by study; patient data included case reports/vignettes, longitudinal prescription and medical claims data, medical records, and synthetic admission notes (Fig. 2B; Table 2). Trial data were primarily drawn from ClinicalTrials.gov, pre-processed trial datasets, and hospital or international databases (**Fig. 2B; Table 3**). Most data provided to model pipelines was unstructured text (Table 2). For articles where LLMs were fine-tuned, ground truth was generally derived from historical enrollments, researcher-determined matches or other LLMs (**Table 2-3**).

Diseases represented in these patient data sets include cancer, stroke, diabetes, cognitive disorders, liver disease, heart failure, or a mix of conditions (**Table 2**). No articles used non-text data like images, and only Unlu et al and Lai et al. used data from ongoing randomized clinical trials. Eleven studies utilized a selection of two or more clinical note types, which could be real or synthetic (**Tables 1-2**). For studies that used real EHR, selecting a subset of note types or from a set of note dates allowed them to select for only the most relevant notes while reducing the inputs to the LLM, increasing efficiency. Sixteen studies used publicly available synthetic or de-identified data (TREC CT 2021-2023[17–19], SIGIR 2016[20], 2018 n2c2[21]), eleven used real patient data, and three used case reports/vignettes (Table 1-2). Notably, Woo et al used Llama-3.1-70B-Instruct to generate eligibility-criteria-based question/answer pairs based on discharge summaries from the publicly-available MIMIC-III data set, while Zihang et al used an LLM to generate simulated answers to patient questionnaires as their patient data inputs.

Rybinski et al used the TREC CT 2021 and 2022 for training and validation, and TREC CT 2023 for testing. Since the latter includes data from earlier years, model performance metrics might be influenced by data leakage.

*Data Pre-Processing*

Data pre-processing pipelines, especially those used for pipelines run on clinical notes, generally included some combination of four stages: enrichment, query generation, retrieval, and restructuring (**Fig. 2C**).

Enrichment refers to the process of augmenting raw text with structured information that enhances semantic interpretability and downstream utility for computational models. In the context of trial matching, enrichment transforms unstructured patient or trial data into a machine-readable representation by extracting key medical concepts, standardizing terminology, and/or clarifying contextual meaning—thereby enabling accurate and efficient alignment between patient profiles and trial eligibility criteria.

While enrichment implementation varied, it generally involved one or more of the following techniques: (1) **Entity extraction**: Identifies clinically meaningful phrases like "Stage IV non-small cell lung cancer" as discrete concepts to enable structured matching. (2) **Concept normalization**: Maps colloquial or alternate terms to standardized medical terminology—for example, recognizing “high blood pressure” as equivalent to “hypertension”—to prevent missed matches due to lexical variation. (3) **Negation detection**: Ensures that the model correctly interprets statements containing negation, such as interpreting “no history of diabetes" as the patient not having diabetes, so the patient is not matched to a diabetes-related trial.

Once enriched, this structured data could be used to generate search queries that retrieve relevant patient or trial segments. Two key query optimization methods were common: (1) **Query expansion**: Adds related terms or synonyms to broaden the search scope (e.g., expanding “lung cancer” to include “NSCLC” or “pulmonary neoplasms”). (2) **Query synthesis**: Repackages the extracted and enriched data into coherent, task-specific prompts or structured queries suitable for the retrieval model.

Retrieval involves extracting relevant portions of patient records based on the generated queries. By reducing the data ultimately input into the LLM, retrieval serves two purposes; (1) it keeps the data size within a given LLM’s context window and (2) it reduces computational costs by using a smaller, less computationally intensive model to reduce the data the larger, more computationally-intensive LLM ultimately needs to process. Retrieval-augmented generation (RAG) is commonly used to identify portions of the queried text most semantically similar to the query; however, this approach can lose chronological context, which is essential to properly analyzing patient data. A variety of approaches could be used to circumvent this issue; for instance, one pipeline combined lexical (BM25) and semantic (MedCPT) retrieval strategies to better preserve clinical chronology.[22] Lexical retrieval served to identify relevant clinical trials by matching exact or near-exact keywords from a set of synthetic data from a given patient to the text in trial eligibility criteria, thereby capturing surface-level correspondences in terminology and phrasing. In contrast, semantic retrieval encoded both patient data and trial descriptions into dense vector representations using pretrained language models, enabling the retrieval of conceptually aligned information even when lexical overlap is limited.

For prompt-based trial matching pipelines, the last data pre-processing step is restructuring. This restructuring generally involves taking the retrieved chunks—usually consisting of one or multiple trial criteria and the corresponding extracted patient information—and reformatting them into a prompt provided to a model to request the model perform patient-criterion, patient-trial, or trial-patient eligibility.

Pre-processed data is then usually fed into one of two matching pipeline types (**Fig. 2D**): (1) **Prompt-based**: Prompts are provided to an LLM, the responses from which are subsequently integrated into article-specific scoring methods to generate ranked outputs. (2) **Embedding-based**: Patient and trial data were embedded in a shared vector space and directly ranked using similarity scores.

LLMs are also applied to tasks beyond eligibility assessment or mapping into a shared vector space—they have been used to convert free text patient or trial data into structured formats like JSONs for use for downstream models; generate synthetic datasets for training smaller models; enrich data sets; and enhance retrieval through query generation, expansion, or synthesis (Fig. 2E). Some models provided rationale for matching decisions or cited specific sentences to support their determination (Fig. 2D).[22–24] This feature not only improved model performance, but increased transparency and the ability of end users to assess accuracy. Depending on the matching type, these pipelines are designed for different end-users — while patient-trial matching systems can generate a list of trials for a for a patient or their care team, trial-patient matching can help review patient data to extract a list of potentially eligible patients for coordinators of a particular trial (Fig. 2E).

*Model Performance & Cost Analysis*

Although comparisons across studies were hindered by heterogeneous tasks and data, intra-study comparisons showed GPT-4 consistently outperformed traditional models for inclusion/exclusion extraction and patient-to-trial matching compared to open-access models, including ones that had been fine-tuned (Table 1). Only a few studies reported on cost and efficiency. In a study by Jin et al, their TrialGPT pipeline reduced clinician screening time by 42.6%.[22] Several studies reported costs associated with LLM-assisted trial matching lower than traditional human screening methods (Table 1). In a study by Gupta et al, a Qwen-1.5-14B-based pipeline, OncoLLM, incurred a per-patient cost of only $0.17 per patient-trial pair, as compared to the cost of $6.18 per pair incurred by GPT-4 (Table 1). For pipelines using either GPT-3.5 or GPT-4, screening cost per-patient ranged between $0.02 and $15.88 per patient-trial pair (Table 1). Of note—in work performed by Unlu et al, use of RAG methods and other strategies dropped that $15.88 value to only 2 cents/patient for GPT-4 (Table 1). For finely-tuned models, training cost is also a consideration; Gupta et al put the cost of training their smaller, fine-tuned model OncoLLM at $2,688. Overall, per-patient processing tended to be fast, with studies reporting between 1 and 12.4 minutes to evaluate a given patient for a trial (Table 1).

## Discussion

There is a pressing need to improve patient screening for patient trials, both to accelerate scientific discovery and to expand access to potentially life-saving treatments. However, the expensive, time-consuming nature of patient matching limits the speed of this process and contributes to limited diversity of patients who ultimately enroll. LLM-based pipelines have the potential to offer a more flexible and scalable alternative to previous rule-based matching systems, reducing the workload for trial staff and facilitating clinical trial success.

One barrier to implementation is that some models struggle with the nuanced and intricate nature of EHR data. Most trial-matching models are trained on data that is synthetic, simplified, abridged, or in structured data formats that, while more easily read by LLMs, do not fully represent the complexities of real-world, long-context EHR data. Their reliance on narrowly defined variables from trial criteria or patient records can limit their generalizability, and real-world applicability. Ideally, trial matching data sets should closely parallel the same real-world data that models will be used with when implemented in the clinic. A key barrier to achieving this goal is the high cost required to collect large, diverse datasets annotated by experts, especially since PHI concerns restrict data sharing. Despite these limitations, the success of models applied to diverse unstructured datasets, such as medical claims data and admission records, demonstrates the versatility of LLMs more broadly—a quality that is likely to increase with ongoing advancements in model architecture and capabilities.

The effectiveness of trial-matching systems hinges not only on model architecture but also on the integration of stakeholder expertise into their design and evaluation. Developing successful pipelines requires collaboration with key stakeholders, such as trial coordinators, who can provide insights into how eligibility criteria are weighted in practice and how patients are evaluated for enrollment. This input can be used both to refine model performance and to ensure that trial-matching tools integrate seamlessly into existing workflows. Currently, GPT-4 offers the strongest performance in trial-matching tasks, with high zero-shot capabilities reducing the need for annotated datasets and advanced retrieval strategies helping to offset computational costs (Fig. 1). For institutions unable to deploy GPT-4 within HIPAA-compliant infrastructure, smaller LLMs present a viable alternative when fine-tuned for specific tasks, offering a balance between accuracy, cost, and data privacy (Fig. 3). Using smaller models can also avoid issues with shifts in model performance and behavior over time, which can be seen with proprietary models like GPT-4.

Standardization of evaluation metrics and evaluation on high-quality public benchmarks is crucial to facilitate comparison of model pipeline performance. For patient-to-criterion matching, metrics such as accuracy, sensitivity, specificity, recall and $F_1$ scores are appropriate. Similar metrics can be used for assessment of patient-to-trial or trial-to-patient matching capabilities, in addition to metrics like NDCG@10, Precision@10,

AUROC and AURPC to assess ranking performance. Accuracy of LLM-generated explanations for eligibility determinations compared to qualified staff should also be determined. Assessing the performance of different pipelines on the same public benchmarks would also facilitate accurate comparisons. However, data memorization by LLMs is also a potential issue when using public data sets, so it is important to also use private data sets for model validation.

Several steps can be taken to mitigate concerns of hallucinations, critical errors and bias. To address concerns of model hallucinations, LLMs can be instructed to provide rationales for their matching decisions, along with explicit citations from the input. These explanations both improve model accuracy and facilitate the correction of model mistakes by human users. Trial staff should also remain a key component of patient assessment, ensuring critical clinical details are not overlooked. Moreover, human-in-the-loop validation studies are necessary to assess whether LLM-based matching pipelines improve human performance or change human behavior. Model performance across diverse patient groups should be assessed, and safeguards should be put in place to prevent models amplifying bias present in training data.

Early trial matching systems show promise in ultimately reducing the time spent by staff in assessing eligibility, as well as the overall cost of trial recruitment. Future work should also assess the cost of data collection and annotation, infrastructure to host the model, personnel to design and maintain it, and the choice of cloud service.

### ***Conclusions***

Recent advances in LLM-based systems offer improved generalizability and accuracy, with models being employed across diverse tasks. Implementation of data pre-processing strategies—such as enrichment, query generation, and retrieval—shows promise for improving performance and reducing computational costs. Intra-study comparisons suggest that GPT-4o achieves the strongest performance in trial-matching tasks, with high zero-shot capabilities reducing reliance on annotated datasets. For institutions unable to deploy GPT-4 within HIPAA-compliant infrastructure, smaller LLMs offer a viable alternative when fine-tuned for specific tasks. Future work should prioritize the use of standardized assessment metrics and test sets to enable cross-study comparison, as well as evaluate cost and bias compared to existing trial matching strategies. To ensure safe and effective implementation, models should be evaluated on real-world data and designed with transparency measures, such as LLM-generated eligibility justifications with sentences cited directly from records. Collaborative development with clinical stakeholders will be essential to ensure these tools augment, rather than replace, human decision-making and integrate effectively into real-world workflows.

## References

1. Wen PY, Weller M, Lee EQ, Alexander BM, Barnholtz-Sloan JS, Barthel FP, et al. Glioblastoma in adults: a Society for Neuro-Oncology (SNO) and European Society of Neuro-Oncology (EANO) consensus review on current management and future directions. Neuro-Oncol. 2020 Aug 17;22(8):1073–113.
2. Yu W, Zhou D, Meng F, Wang J, Wang B, Qiang J, et al. The global, regional burden of pancreatic cancer and its attributable risk factors from 1990 to 2021. BMC Cancer. 2025 Jan 31;25(1):186.
3. Mani K, Deng D, Lin C, Wang M, Hsu ML, Zaorsky NG. Causes of death among people living with metastatic cancer. Nat Commun. 2024 Feb 19;15(1):1519.
4. Taylor K, Francesca P, Cru MJ, Ronte H, Haughey J. Intelligent clinical trials [Internet]. [cited 2025 Jan 26]. Available from: https://www2.deloitte.com/content/dam/insights/us/articles/22934_intelligent-clinical-trials/DI_Intelligent-clinical-trials.pdf
5. Kasenda B, von Elm E, You J, Blümle A, Tomonaga Y, Saccilotto R, et al. Prevalence, Characteristics, and Publication of Discontinued Randomized Trials. JAMA. 2014 Mar 12;311(10):1045–52.
6. Wong AR, Sun V, George K, Liu J, Padam S, Chen BA, et al. Barriers to Participation in Therapeutic Clinical Trials as Perceived by Community Oncologists. JCO Oncol Pract. 2020 Sept;16(9):e849–58.
7. Durden K, Hurley P, Butler DL, Farner A, Shriver SP, Fleury ME. Provider motivations and barriers to cancer clinical trial screening, referral, and operations: Findings from a survey. Cancer. 2024;130(1):68–76.
8. Penberthy LT, Dahman BA, Petkov VI, DeShazo JP. Effort Required in Eligibility Screening for Clinical Trials. J Oncol Pract. 2012 Nov;8(6):365–70.
9. Guerra CE, Fleury ME, Byatt LP, Lian T, Pierce L. Strategies to Advance Equity in Cancer Clinical Trials. Am Soc Clin Oncol Educ Book Am Soc Clin Oncol Annu Meet. 2022 Apr;42:1–11.
10. Yuan C, Ryan PB, Ta C, Guo Y, Li Z, Hardin J, et al. Criteria2Query: a natural language interface to clinical databases for cohort definition. J Am Med Inform Assoc. 2019 Apr 1;26(4):294–305.
11. Thadani SR, Weng C, Bigger JT, Ennever JF, Wajngurt D. Electronic Screening Improves Efficiency in Clinical Trial Recruitment. J Am Med Inform Assoc JAMIA. 2009;16(6):869–73.
12. Shriver SP, Arafat W, Potteiger C, Butler DL, Beg MS, Hullings M, et al. Feasibility of institution-agnostic, EHR-integrated regional clinical trial matching. Cancer. 2024;130(1):60–7.
13. Haddad TC, Helgeson J, Pomerleau K, Makey M, Lombardo P, Coverdill S, et al. Impact of a cognitive computing clinical trial matching system in an ambulatory oncology practice. J Clin Oncol. 2018 May 20;36(15_suppl):6550–6550.
14. Gupta S, Basu A, Nievas M, Thomas J, Wolfrath N, Ramamurthi A, et al. PRISM: Patient Records Interpretation for Semantic clinical trial Matching system using large language models. Npj Digit Med. 2024 Oct 28;7(1):1–12.
15. Shi H, Zhang J, Zhang K. Enhancing Clinical Trial Patient Matching through Knowledge Augmentation with Multi-Agents [Internet]. arXiv; 2024 [cited 2025 Jan 19]. Available from: http://arxiv.org/abs/2411.14637
16. Unlu O, Shin J, Mailly CJ, Oates MF, Tucci MR, Varugheese M, et al. Retrieval-Augmented Generation–Enabled GPT-4 for Clinical Trial Screening. NEJM AI. 2024 June 27;1(7):AIoa2400181.
17. 2021 TREC Clinical Trials Track [Internet]. [cited 2025 Jan 15]. Available from: http://www.trec-cds.org/2021.html#documents
18. 2022 TREC Clinical Trials Track [Internet]. [cited 2025 Jan 15]. Available from: https://www.trec-cds.org/2023.html
19. 2023 TREC Clinical Trials Track [Internet]. [cited 2025 Jan 15]. Available from: https://www.trec-cds.org/2023.html
20. Koopman B, Zuccon G. A Test Collection for Matching Patients to Clinical Trials. In: Proceedings of the 39th International ACM SIGIR conference on Research and Development in Information Retrieval [Internet]. Pisa Italy: ACM; 2016 [cited 2025 Jan 4]. p. 669–72. Available from: https://dl.acm.org/doi/10.1145/2911451.2914672
21. Stubbs A, Filannino M, Soysal E, Henry S, Uzuner Ö. Cohort selection for clinical trials: n2c2 2018 shared task track 1. J Am Med Inform Assoc. 2019 Nov 1;26(11):1163–71.
22. Jin Q, Wang Z, Floudas CS, Chen F, Gong C, Bracken-Clarke D, et al. Matching patients to clinical trials with large language models. Nat Commun. 2024 Nov 18;15(1):9074.
23. Nievas M, Basu A, Wang Y, Singh H. Distilling large language models for matching patients to clinical trials. J Am Med Inform Assoc JAMIA. 2024 Sept 1;31(9):1953–63.
24. Wornow M, Lozano A, Dash D, Jindal J, Mahaffey KW, Shah NH. Zero-Shot Clinical Trial Patient Matching with LLMs. NEJM AI. 2024 Dec 24;0(0):AIcs2400360.

25. Covidence - Better systematic review management [Internet]. Covidence. [cited 2025 Sept 4]. Available from: https://www.covidence.org/
26. Beattie J, Neufeld S, Yang D, Chukwuma C, Gul A, Desai N, et al. Utilizing Large Language Models for Enhanced Clinical Trial Matching: A Study on Automation in Patient Screening. Cureus. 2024 May;16(5):e60044.
27. Beattie J, Owens D, Navar AM, Giuliani Schmitt L, Taing K, Neufeld S, et al. ChatGPT augmented clinical trial screening. Mach Learn Health. 2025 July;1(1):015005.
28. Cerami E, Trukhanov P, Paul MA, Hassett MJ, Riaz IB, Lindsay J, et al. MatchMiner-AI: An Open-Source Solution for Cancer Clinical Trial Matching [Internet]. arXiv; 2024 [cited 2025 Jan 11]. Available from: http://arxiv.org/abs/2412.17228
29. Chowdhury S, Rajaganapathy S, Yu Y, Tao C, Vassilaki M, Zong N. Matching Patients to Clinical Trials using LLaMA 2 Embeddings and Siamese Neural Network. medRxiv. 2024 June 30;2024.06.28.24309677.
30. Datta S, Lee K, Huang LC, Paek H, Gildersleeve R, Gold J, et al. Patient2Trial: From Patient to Participant in Clinical Trials Using Large Language Models. Inform Med Unlocked. 2025 Jan 17;101615.
31. Devi A, Uttrani S, Singla A, Jha S, Dasgupta N, Natarajan S, et al. Quantitative Analysis of GPT-4 model: Optimizing Patient Eligibility Classification for Clinical Trials and Reducing Expert Judgment Dependency. In: Proceedings of the 2024 8th International Conference on Medical and Health Informatics [Internet]. New York, NY, USA: Association for Computing Machinery; 2024 [cited 2024 Dec 23]. p. 230–7. (ICMHI '24). Available from: https://dl.acm.org/doi/10.1145/3673971.3674014
32. Devi A, Uttrani S, Singla A, Jha S, Dasgupta N, Natarajan S, et al. Automating Clinical Trial Eligibility Screening: Quantitative Analysis of GPT Models versus Human Expertise. In: Proceedings of the 17th International Conference on PErvasive Technologies Related to Assistive Environments [Internet]. New York, NY, USA: Association for Computing Machinery; 2024 [cited 2024 Dec 22]. p. 626–32. (PETRA '24). Available from: https://dl.acm.org/doi/10.1145/3652037.3663922
33. Ferber D, Hilgers L, Wiest IC, Leßmann ME, Clusmann J, Neidlinger P, et al. End-To-End Clinical Trial Matching with Large Language Models [Internet]. arXiv; 2024 [cited 2025 Jan 19]. Available from: http://arxiv.org/abs/2407.13463
34. Gueguen L, Olgiati L, Brutti-Mairesse C, Sans A, Le Texier V, Verlingue L. A prospective pragmatic evaluation of automatic trial matching tools in a molecular tumor board. Npj Precis Oncol. 2025 Jan 27;9(1):28.
35. Gui X, Lv H, Wang X, Lv L, Xiao Y, Wang L. Enhancing hepatopathy clinical trial efficiency: a secure, large language model-powered pre-screening pipeline. BioData Min. 2025 June 14;18:42.
36. Jullien M, Bogatu A, Unsworth H, Freitas A. Controlled LLM-based Reasoning for Clinical Trial Retrieval [Internet]. arXiv; 2024 [cited 2025 Jan 19]. Available from: http://arxiv.org/abs/2409.18998
37. Kusa W, Mendoza ÓE, Knoth P, Pasi G, Hanbury A. Effective matching of patients to clinical trials using entity extraction and neural re-ranking. J Biomed Inform. 2023 Aug 1;144:104444.
38. Kusa W, Styll P, Seeliger M, Espitia Mendoza Ó, Hanbury A. DoSSIER at TREC 2023 Clinical Trials Track. In: # PLACEHOLDER_PARENT_METADATA_VALUE# [Internet]. NIST; 2023 [cited 2025 Jan 19]. Available from: https://repositum.tuwien.at/handle/20.500.12708/203878
39. Lai SM, Malik AM, Sathe TS, Silvestri CJ, Manji GA, Kluger MD. A Proof-of-Concept Large Language Model Application to Support Clinical Trial Screening in Surgical Oncology [Internet]. medRxiv; 2024 [cited 2025 Feb 21]. p. 2024.09.20.24314053. Available from: https://www.medrxiv.org/content/10.1101/2024.09.20.24314053v2
40. Lin J, Xu H, Wang Z, Wang S, Sun J. Panacea: A foundation model for clinical trial search, summarization, design, and recruitment [Internet]. medRxiv; 2024 [cited 2024 Dec 23]. p. 2024.06.26.24309548. Available from: https://www.medrxiv.org/content/10.1101/2024.06.26.24309548v1
41. Peikos G, Symeonidis S, Kasela P, Pasi G. Utilizing ChatGPT to Enhance Clinical Trial Enrollment [Internet]. arXiv; 2023 [cited 2025 Jan 18]. Available from: http://arxiv.org/abs/2306.02077
42. Peikos G, Kasela P, Pasi G. Leveraging Large Language Models for Medical Information Extraction and Query Generation [Internet]. arXiv; 2024 [cited 2025 Aug 27]. Available from: http://arxiv.org/abs/2410.23851
43. Rahmanian M, Fakhrahmad SM, Mousavi SZ. Towards Efficient Patient Recruitment for Clinical Trials: Application of a Prompt-Based Learning Model [Internet]. arXiv.org. 2024 [cited 2025 Feb 1]. Available from: https://arxiv.org/abs/2404.16198v1

44. Ruan J, Su Q, Chen Z, Huang J, Li Y. CPRS: a clinical protocol recommendation system based on LLMs. Int J Med Inf. 2024 Dec 4;195:105746.
45. Rybinski M, Kusa W, Karimi S, Hanbury A. Learning to match patients to clinical trials using large language models. J Biomed Inform. 2024 Nov 1;159:104734.
46. Unlu O, Shin J, Mailly CJ, Oates MF, Tucci MR, Varugheese M, et al. Retrieval Augmented Generation Enabled Generative Pre-Trained Transformer 4 (GPT-4) Performance for Clinical Trial Screening. MedRxiv Prepr Serv Health Sci. 2024 Feb 8;2024.02.08.24302376.
47. Wong C, Zhang S, Gu Y, Moung C, Abel J, Usuyama N, et al. Scaling Clinical Trial Matching Using Large Language Models: A Case Study in Oncology [Internet]. arXiv; 2023 [cited 2025 Jan 11]. Available from: http://arxiv.org/abs/2308.02180
48. Woo EG, Burkhart MC, Alsentzer E, Beaulieu-Jones BK. Synthetic data distillation enables the extraction of clinical information at scale. Npj Digit Med. 2025 May 10;8(1):267.
49. Yuan J, Tang R, Jiang X, Hu X. Large Language Models for Healthcare Data Augmentation: An Example on Patient-Trial Matching. AMIA Annu Symp Proc. 2024 Jan 11;2023:1324–33.
50. Zhuang S, Koopman B, Zuccon G. Team IELAB at TREC Clinical Trial Track 2023: Enhancing Clinical Trial Retrieval with Neural Rankers and Large Language Models [Internet]. arXiv; 2024 [cited 2024 Dec 21]. Available from: http://arxiv.org/abs/2401.01566
51. ZiHang C, QianMin S, GaoYi C, JiHan H, Ying L. Enhanced Pre-Recruitment Framework for Clinical Trial Questionnaires Through the Integration of Large Language Models and Knowledge Graphs [Internet]. Rochester, NY: Social Science Research Network; 2024 [cited 2025 Jan 19]. Available from: https://papers.ssrn.com/abstract=4713177
52. A S, Ö U. Annotating longitudinal clinical narratives for de-identification: The 2014 i2b2/UTHealth corpus. J Biomed Inform [Internet]. 2015 Dec [cited 2025 Aug 27];58 Suppl(Suppl). Available from: https://pubmed.ncbi.nlm.nih.gov/26319540/
53. A S, Ö U. Annotating risk factors for heart disease in clinical narratives for diabetic patients. J Biomed Inform [Internet]. 2015 Dec [cited 2025 Aug 27];58 Suppl(Suppl). Available from: https://pubmed.ncbi.nlm.nih.gov/26004790/
54. Beattie J, Owens D, Navar AM, Schmitt LG, Taing K, Neufeld S, et al. Large Language Model Augmented Clinical Trial Screening [Internet]. medRxiv; 2024 [cited 2024 Dec 22]. p. 2024.08.27.24312646. Available from: https://www.medrxiv.org/content/10.1101/2024.08.27.24312646v1
55. Voorhees EM, Hersh W. Overview of the TREC 2012 Medical Records Track. NIST [Internet]. 2013 June 28 [cited 2025 Aug 27]; Available from: https://www.nist.gov/publications/overview-trec-2012-medical-records-track
56. Woo E, Burkhart MC, Alsentzer E, Beaulieu-Jones B. MIMIC-III-Ext-Synthetic-Clinical-Trial-Questions [Internet]. PhysioNet; [cited 2025 Sept 2]. Available from: https://physionet.org/content/mimic-ext-synth-trial-question/1.0.0/
57. Woo E, Burkhart MC, Alsentzer E, Beaulieu-Jones B. MIMIC-IV-Ext-Apixaban-Trial-Criteria-Questions [Internet]. PhysioNet; [cited 2025 Sept 2]. Available from: https://physionet.org/content/mimic-iv-ext-apixaban-trial/1.0.0/
58. Yang Y, Jayaraj S, Ludmir E, Roberts K. Text Classification of Cancer Clinical Trial Eligibility Criteria. AMIA Annu Symp Proc. 2024 Jan 11;2023:1304–13.

## Methods

*Article Identification*

We identified 126 unique citations across three academic databases and one preprint server using a structured keyword search combining trial-related and LLM-related terms between January 1st , 2020 and March 1st, 2025 (Fig. 1). Three articles were published as pre-prints within this time frame, then published in peer-reviewed journals after March 1st, 2025, reflecting a common trend of first-step publishing in pre-print servers. The search string included: ("match[tiab]" OR "screen*[tiab]") AND ("clinical trial"[tiab] OR "clinical trials"[tiab]), combined with ("large language models"[tiab] OR "LLMs"[tiab] OR "large language model"[tiab] OR "LLM"[tiab] OR "ChatGPT"[tiab] OR "GPT-4"[tiab] OR "LLAMA"[tiab] OR "GPT-3.5"[tiab] OR "GPT-4"[tiab]). Covidence, a web-based review management tool, was employed to organize references, identify duplicate records, and document study inclusion/exclusion decisions.[25] After excluding 51 duplicates and 91 ineligible records, 31 full-text articles were included (Fig. 1). All remaining records were accessible via full-text retrieval. Four were published in 2023, 23 in 2024, and four in 2025, reflecting rapid developments in the field.

**Data availability**
No new data were generated or analyzed in support of this review. Data sharing is not applicable.

**Code availability**
All data supporting this review are available within the cited articles. No new data were created.

**Competing interests statement**

The authors declare no competing interests.

## Tables

| Table 1. Model Performance Assessment | | | | | | | |
|---|---|---|---|---|---|---|---|
| **Study** | **Study Data Set** | **Main Model(s)** | **Patient-to-Criterion Metrics** | **Patient-to-Trial Metrics** | **Trial-to-Patient Metrics** | **Model Comparisons†** | **Cost/Efficiency** |
| Beattie et al., 2024[26] | - 2018 n2c2 | GPT-3.5 Turbo; GPT-4 | GPT-3.5 Turbo:<br>- Accuracy: 0.81<br>- Sensitivity: 0.80<br>- Specificity: 0.82<br>- Micro $F_1$: 0.79<br><br>GPT-4:<br>- Accuracy: 0.87<br>- Sensitivity: 0.85<br>- Specificity: 0.89<br>- Micro $F_1$: 0.86 | N/A | N/A | N/A | Model-processing time: 1-5 minutes per patient for all included criteria |
| Beattie et al., 2024[27] | - Article-specific patient and trial data | GPT-3.5; GPT-4 | GPT-3.5:<br>- Accuracy: 0.761<br>- Sensitivity: 0.776<br>- Specificity: 0.732<br>- Youden Index: 0.357<br><br>GPT-4:<br>- Accuracy: 0.838<br>- Sensitivity: 0.839<br>- Specificity: 0.830<br>- Youden Index: 0.668 | GPT-3.5:<br>- Median AUC: 0.64<br>- Median accuracy: 0.65<br>-Median sensitivity: 0.68<br>- Median specificity: 0.69<br>- Median Youden index: 0.27<br><br>GPT-4<br>- Median AUC: 0.74<br>- Median accuracy: 0.70<br>- Median sensitivity: 0.66<br>- Median specificity: 0.74 | N/A | Proprietary: GPT-3.5; GPT-4 | Screening cost per patient:<br>- GPT-3.5: $0.02-$0.03<br>- GPT-4: $0.15-$0.27.<br><br>Screening time per patient:<br>- GPT-3.5: 1.4-3 minutes<br>- GPT-4: 7.9-12.4 minutes |
| Cerami et al., 2024[28] | - Article-specific patient and trial data | TrialSpace text embedding + TrialChecker classification model (fine-tuned RoBERTa-Large; open-access) | N/A | DFCI trial enrollment dataset:<br>- Precision@10: 0.89<br>- MAP@10: 0.93<br><br>DFCI standard-of-care dataset:<br>- Precision@10: 0.87<br>- MAP@10: 0.91 | DFCI trial enrollment dataset:<br>- Precision@20: 0.91<br>- MAP@20: 0.93<br><br>DFCI standard-of-care dataset:<br>- Precision@20: 0.87<br>- MAP@20: 0.90 | N/A | N/A |
| Chowdhury et al., 2024[29] | - Article-specific patient and trial data | Siamese-Patient-Trial-Matching with LLaMA 2 to initialize embeddings of inputs (open-access) | N/A | Siamese-PTM (fine-grained): $F_1$ score of 0.92<br>*Note: article describes binary classification task of eligible/not eligible so could be labeled either “patient-trial” or “trial-patient” matching | | N/A | N/A |
| Datta et al., 2025[30] | - TREC 2023 CT | GPT-4 | N/A | - Precision@10: 0.7351<br>- NDCG@10: 0.8109 | N/A | N/A | N/A |
| Devi et al., 2024[32] | - Article-specific patient and trial data | GPT-3.5 Turbo | N/A | N/A | Accuracy of GPT-3.5:<br>No tuning tested on 20 data points: 100%<br>Tuned and tested on 20 data points: 95%<br>No tuning tested on 100 data points: 79%<br>Tuned and tested on 50 data points: 82% | N/A | N/A |
| Devi et al., 2024[31] | - Article-specific patient and trial data | GPT-4 | N/A | N/A | Accuracy:<br>No tuning, tested on 20 data points: 95%<br>Tuned and tested on 20 data points: 100%<br>No tuning, tested on 100 data points: 86% | N/A | N/A |
| Ferber et al., 2024[33] | - Article-specific patient and trial data | GPT-4o | Accuracy: 88.0% | For 14/15 patients, their “target trial” was listed within the top 15 trial matches identified by the algorithm | N/A | N/A | N/A |
| Gueguen et al, 2025[34] | TrialGPT to re-rank outputs from DigitalECMT | TrialGPT with Qwen2.5-7B-Instruct as the LLM | N/A | Use of LLM-based re-ranking on results of DigitalECMT increased NDCG@3 from 0.61 to 0.64 | N/A | N/A | N/A |
| Gui et al, 2025[35] | - Article-specific patient and trial data | - Anthropo-morphized Experts’ Chain of Thought<br>- LLMs converted eligibility criteria into questions: GPT-4o, Google Gemini Advanced, Anthropic Claude 3.5 Sonnet | Pathway A, Majority Vote<br>- Precision: ~0.921<br>- Recall: ~0.82 | Pathway A, Majority Vote<br>- Precision: ~0.922<br>- Recall: ~0.819<br>*Note: article describes binary classification task of eligible/not eligible so could be labeled either “patient-trial” or “trial-patient” matching | | - QWEN1.5, BAICHUAN, and GLM | Pathway A, Majority Vote<br>- Efficiency: ~0.44 seconds/task |
| Gupta et al., 2024[14] | - Article-specific patient and trial data | OncoLLM (fine-tuned Qwen-1.5 14B model; open-access) | Accuracy:<br>- OncoLLM: 63%<br>- GPT-3.5 Turbo: 53%<br>- GPT-4: 68%<br>- Qwen14B-Chat: 43%<br>- Mitral-7B-Instruct: 41%<br>- Mixtral-8 X 7B-Instruct: 49%<br>- Meditron: 51%<br>- MedLlama: 55%<br>- TrialLlama: 57% | Percentage of time model ranks ground truth trials in the top-3 positions among 10 considered trials<br>- OncoLLM: 65%<br>- GPT-3.5 Turbo (iterative): 61% | Normalized Discounted Cumulative Gain<br>- OncoLLM: 68%<br>- GPT-3.5 Turbo: 62% | - Proprietary: GPT-3.5 Turbo and GPT-4<br><br>- Open-source: Qwen14B-Chat, Mistral-7B-Instruct, Mixtral-8 × 7B-Instruct, Meditron, MedLlama, and TrialLlama | Operating cost:<br>- OncoLLM: ~$170<br>- GPT-4: ~ $6055<br><br>Cost of a single patient-trial match:<br>- OncoLLM: ~$0.17 per patient-trial pair<br>- GPT-4: ~ $6.18 per patient-trial pair |
| Jin et al., 2024[22] | - SIGIR 2016<br>- TREC 2021 CT<br>- TREC 2022 CT | TrialGPT (GPT-4) (zero-shot) | TrialGPT-Matching:<br>- Accuracy: 87.3% | TrialGPT-Ranking:<br>- NDCG@10: 0.7252<br>- Precision@10: 0.6724 | N/A | - Proprietary: GPT-3.5<br><br>- Open-source (for trial ranking): SciFive, BioBERT, PubMedBERT, SapBERT, and BioLinkBERT | Reduction in screening time for patient recruitment: 42.6% |
| Jullien et al., 2024[36] | - TREC 2022 CT | GPT-4 Turbo (proprietary) | N/A | - NDCG@10: 0.679<br>- Precision@10: 0.730<br>- Precision@25: 0.630<br>- MRR: 0.860 | N/A | - GPT-3.5 (proprietary)<br>- TREC SOTA (open-source)<br>*BM-25 for initial ranking | N/A |
| Kusa et al., 2023[37] | - TREC 2021 CT<br>- TREC 2022 CT | TCRR with BioBERT (open-access) | N/A | - NDCG@5: 0.627<br>- NDCG@10: 0.604<br>- Precision@10: 0.482<br>- Reciprocal Rank: 0.672 | N/A | - Open-source: MonoBERT, TraditionalRR, TCRR initialized with *bert-base-uncased*, BioBERT, ClinicalBERT | N/A |
| Kusa et al., 2023[38] | - TREC 2023 CT with structured data converted to unstructured text using GPT-3.5 | TCRR neural re-ranking model with BlueBERT (open-access) and GPT-3.5 (zero-shot) to refine results | N/A | DoSSIER_3:<br>- nDCG@5: 0.6653<br>- nDCG@10: 0.6837<br>- Precision@10: 0.5838<br>- Reciprocal Rank: 0.6421 | N/A | N/A | N/A |
| Lai et al., 2024[39] | - Article-specific patient and trial data | GPT-4o (zero-shot) | GPT-4o:<br>- Accuracy: 96.7% in agreeing with human raters on binary eligibility criteria | GPT-4o:<br>- Accuracy: 90.7% of eligible patient-trial matches<br>- Sensitivity: 87.5-100% for 8/9 trials<br>- Specificity: 73.3-100% for all 9 trials | N/A | N/A | - Median cost to screen a single patient: $0.67 (range: $0.63-$0.74)<br>- Median time elapsed per patient: 138 seconds (range: 130-146)<br>- Median total token usage: 112,266.5 tokens (range: 102982.0-122174.2) |
| Lin et al., 2024[40] | - TREC 2021 CT<br>- SIGIR 2016<br>- TrialAlign | Panacea (finely-tuned Mistral-7B-Base model27; open-access) | Yes; no metrics provided | F1, precision, and recall | N/A | - Open-source: BioMistral, Mistral, Zephyr, LLAMA-2, MedAlpaca, Meditron | N/A |
| Nievas et al., 2024[23] | - SIGIR 2016<br>- TREC CT 2021<br>- TREC CT 2022 | Trial-LLAMA 70B (open-access) | Trial-LLAMA 70B:<br>- Implicit CLA: 68.77<br>- Explicit CLA: 59.9<br><br>GPT-4:<br>- Implicit CLA: 75.31<br>- Explicit CLA: 58.8 | Trial-LLAMA 70B:<br>- NDCG@10: 0.6636<br>- Precision@10: 0.5886<br>- AUROC: 0.6528<br>- AURPC: 0.6515<br><br>GPT-4:<br>- NDCG@10: 0.7728<br>- Precision@10: 0.7005<br>- AUROC: 0.7390<br>- AURPC: 0.7038 | N/A | - Proprietary: GPT-3.5 and GPT-4<br><br>- Open-source: LLAMA 7B, LLAMA 13B, and LLAMA 70B | N/A |
| Peikos et al., 2023[41] | - TREC 2021<br>- TREC 2022 | GPT-3.5 Turbo | N/A | TREC 2021<br>- R-precision: 0.212<br>- Binary preference: 0.275<br>- Precision@10: 0.323<br>- P- recision @25: 0.261<br>- MRR: 0.541<br>- nDCG@10: 0.512<br><br>TREC 2022<br>- R-precision: 0.276<br>- Binary preference: 0.298<br>- Precision @10: 0.372<br>- Precision @25: 0.338<br>- MRR: 0.576<br>- nDCG@10: 0.517 | N/A | N/A | N/A |
| Peikos et al, 2024[42] | - TREC 2021<br>- TREC 2022 | GPT-3.5 Turbo | N/A | TREC 2021**<br><br>GPT-3.5:<br>- NDCG@10: 0.486<br>- Binary preference: 0.213<br>- Precision@10: 0.276<br>- Recall@25: 0.115<br>- MRR: 0.440<br><br>Qwen2-7B-Instruct:<br>- NDCG@10: 0.476<br>- Binary preference: 0.216<br>- Precision@10: 0.285<br>- Recall@25: 0.107<br>- MRR: 0.465 | N/A | -Proprietary: GPT-4<br><br>-Open access: Qwen2; Phi3-medium-4k-Instruct; Phi3-mini-4k-Instruct; Medical-Llama3-8B | N/A |
| Rahmanian et al, 2024[43] | - n2c2 2018 | GPT-3.5 Turbo | Overall (micro):<br>- F1: 0.9061<br>- AUC: 0.9035<br><br>Overall (macro):<br>- F1: 0.8060<br>- AUC: 0.7949 | N/A | N/A | N/A | N/A |
| Ruan et al., 2024[44] | - Article-specific patient and trial data | GPT-4 with a knowledge graph | N/A | GPT-4:<br>- MSE: 27.27<br>- RMSE: 5.22<br>*Measures similarity to criteria of another trial identified by the patient | N/A | - Open-source: 11 models from the SBERT family | N/A |
| Rybinski et al., 2024[45] | - TREC 2021 CT<br>- TREC 2022 CT<br>- TREC 2023 CT | Fine-tuned GPT-3.5 Turbo with BM25 retrieval (zero-shot for final eligibility determination) | N/A | BM25 with TCRR and GPT-3.5 Turbo re-ranking:<br>- nDCG@1000: 0.375<br>- nDCG@10: 0.777<br>- Precision@10: 0.697<br>- Reciprocal Rank: 0.783<br><br>BM25 with chain-of-thought GPT-4o:<br>- nDCG@1000: 0.504<br>- nDCG@10: 0.785<br>- Precision@10: 0.603<br>- Reciprocal Rank: 0.844 | N/A | - Proprietary: GPT4-o | - Cost of re-ranking with GPT-3.5 Turbo: ~$0.25 per 100 API calls (per patient) |
| Shi et al., 2024[15] | - 2018 n2c2<br>- Synthesized a trial for which 28 patients from the 2018 n2c2 cohort would meet all criteria | MAKA (proprietary) | MAKA:<br>- Accuracy: 0.909<br>- Precision: 0.822<br>- Recall: 0.846<br>- $F_1$-score: 0.828<br><br>Strategy by Wornow et al.:<br>- Accuracy: 0.884<br>- Precision: 0.727<br>- Recall: 0.0.894<br>- $F_1$-score: 0.785 | N/A | MAKA:<br>- Accuracy: 0.9306<br>- Precision: 0.6333<br>- Recall: 0.6786<br>- $F_1$-score: 0.6552 | - Strategies employed by Wornow et al and Beattie et al | N/A |
| Unlu et al., 2024[46] | - Article-specific patient and trial data | RAG-Enabled Clinical Trial Infrastructure for Inclusion Exclusion Review (RECTIFIER) – using GPT-4 Vision | RECTIFIER:<br>- Sensitivity: 75-100%<br>- Specificity: 92.1-100%<br>- PPV: 75-100%<br>- MCC: 97.9-100%<br><br>Study Staff:<br>- Sensitivity: 66.7-100%<br>- Specificity: 82.1-100%<br>- PPV: 50-100%<br>- MCC: 91.7-100% | RECTIFIER<br>- Sensitivity: 92.3%<br>- Specificity: 93.9%<br>- PPV: 98.1%<br>- NPV: 78.6%<br>- Accuracy: 92.7%<br>- MCC: 81.3%<br><br>Study staff<br>- Sensitivity: 90.8%<br>- Specificity: 83.6%<br>- PPV: 94.9%<br>- NPV: 73.0%<br>- Accuracy: 89.1%<br>- MCC: 71.1% | N/A | - Proprietary: GPT-3.5 | RECTIFIER:<br>- Individual-question approach: average of 11 cents/patient<br>- Combined-question approach: average of 2 cents/patient<br><br>Cost without RAG:<br>- GPT-4: $15.88 per patient<br>- GPT-3.5: $1.59 per patient |
| Wong et al, 2023[47] | - Article-specific patient and trial data | GPT-4 (3-shot) | Performed but metrics not provided | GPT-3.5 (zero-shot)<br>Precision: 88.5<br>Recall: 11.6<br>F1: 20.6<br><br>GPT-4 (zero-shot)<br>Precision: 86.7<br>Recall: 46.8<br>F1: 60.8<br><br>GPT-4 (3-shot)<br>Precision: 87.6<br>Recall: 67.3<br>F1: 76.1<br>*Note: article describes binary classification task of eligible/not eligible so could be labeled either "patient-trial" or "trial-patient" matching | | - Proprietary: GPT-3.5 | N/A |
| Woo, 2024[48] | - 2018 n2c2<br>- Article-specific patient and trial data | Llama-3.1-8B-All | Llama-3.1-8B-All on MIMIC-IV-based data set:<br>Balanced accuracy: 0.93<br>Micro-F1: 0.94 | N/A | N/A | - Open-access: Llama-3.1-8B, Llama3.1-70B | Cost of evaluating the Apixaban criteria for 10,000.patients:<br>- Llama-3.1-8B: $929<br>- Llama3.1-70B: $4066 |
| Wornow et al., 2024[24] | - 2018 n2c2<br>- 1 novel exclusion criterion from a pulmonary arterial hypertension clinical trial<br>- SIGIR 2016 | Zero-shot GPT-4 with ACIN prompting strategy | - Precision: 0.91<br>- Recall: 0.92<br>- Macro-$F_1$: 0.81<br>- Micro-$F_1$: 0.93 | N/A | N/A | - Proprietary: GPT-3.5<br>- Open-source: Llama-2-70b, Mixtral-8x7B, Qwen2-72b, and Llama-3-70b | - Cost to screen a single patient: ~$1.55<br>- Evaluation time per patient: ~1 minute |
| Yuan et al., 2024[49] | - Article-specific patient and trial data | LLM-based patient-trial matching with GPT-4 (LLM-PTM) | - Precision: 0.964<br>- Recall: 0.862<br>- $F_1$ Score: 0.910 | - Precision: 0.801<br>- Recall: 0.830<br>- $F_1$ Score: 0.815 | N/A | N/A | N/A |
| Zhuang et al., 2024[50] | - TREC 2022 CT<br>- TREC 2023 CT | Hybrid PubmedBERT-based retriever with GPT-4 re-ranking | N/A | Bi-encoder dense retriever:<br>NDCG@10: 0.5768<br>P@10: 0.3243<br>Recall@1000: 0.3670<br><br>PLADEv2 sparse retriever:<br>NDCG@10: 0.5971<br>P@10: 0.3243<br>Recall@1000: 0.3482<br><br>Hybrid:<br>- NDCG@10: 0.5763<br>- P@10: 0.2946<br>- Recall@1000: 0.3878<br><br>CE_weighted:<br>- NDCG@10: 0.6716<br>- P@10: 0.4432<br>- Recall@1000: 0.3878<br><br>GPT-4:<br>- NDCG@10: 0.7363<br>- P@10: 0.5108<br>- Recall@1000: 0.3878 | N/A | GPT-3.5 Turbo (proprietary) to generate extra training data | N/A |
| Zihang et al., 2025[51] | - Article-specific patient and trial data | llama3-70b-instruct | Performed but metrics not provided | Accuracy:<br>- GLM-3-Turbo: 0.8973<br><br>- GLM-4: 0.9139<br><br>- llama3-70b-instruct: 0.9285<br><br>- Qwen-Turbo: 0.9166 | N/A | GLM-3-Turbo, GLM-4, Qwen-Turbo | N/A |

* 2018 n2c2 denotes 2018 National Natural Language Processing Clinical Challenges cohort; ACIN, All criteria, individual notes: All notes are merged into a single prompt, but the model assesses one criterion at a time, requiring reprompting for each criterion; CLA: criterion-level accuracy; MAKA, Multi-Agents for Knowledge Augmentation; MCC = Mathew correlation coefficient; MRR = mean reciprocal rank; NCDG@10: Normalized Cumulative Discounted Gain; NDCG@10: Normalized Discounted Cumulative Gain; NPV: negative predictive value; PPV: positive predictive value; Program-rather-than-prompt: Ensures responses adhere to required format using structured programming objects instead of free-text prompts; QGMT, Query Generation, Medical Role & Task Description: Provides contextual information to ChatGPT and instructs it to generate a single keyword-based query; RAG, Retrieval-Augmented Generation; SIGIR 2016, Special

Interest Group on Information Retrieval 2016 cohort; TCRR, Topical and Criteria Re-Ranking involves a two-step training schema focusing on topical relevance and eligibility classification; TREC 2021/2022/2023 CT, Text Retrieval Conference Clinical Trials Track.

† Models compared within a given study include any models that were assessed by the authors alongside their best-performing LLM.

**Additional metrics were calculated for TREC 2022 due to spacing concerns.

Source: Data compiled from the studies included in this systematic review.

**Table 2. Patient Data Used for Training and Testing of LLMs**

| Study Dataset | Source of Data | Data Description | Data Size | Patient Characteristics | Data Type | Publicly Available? |
|---|---|---|---|---|---|---|
| 2018 n2c2[21] | 2014 i2b2/UTHealth shared tasks[52,53] | Real, de-identified clinical notes; synthetic clinical trial | 288 patients (2-5 records/patient); 13 predefined inclusion criteria | Diabetes | Unstructured text | Yes |
| Beattie et al., 2024[54] | Enrollment list of phase II trial investigating hypo fractionated radiation therapy for head and neck cancer; head and neck radiation oncology team | Real patient notes from surgical oncology, radiation oncology and medical oncology from the last 6 months; relevant pathology reports, imaging reports and lab results going back up to 1 year | 35 patients enrolled in trial; 40 randomly identified patients seen by head and neck radiation oncology team | Head and neck cancer | Unstructured text | Not Reported |
| Cerami et al., 2024[28] | Dana-Farber Cancer Institute (DFCI) Oncology Data Retrieval System | - Retrospective trial enrollment data set: Real EHR data (all unstructured clinical notes, imaging reports, and pathology reports) for all adults who enrolled in cancer treatment clinical trials at DFCI from January 2016 to April 2024<br>- Standard of care (SOC) treatment dataset: Real data for patients who started SOC systemic therapies at DFCI from 2016 to 2024 | - Retrospective trial enrollment data set: 16,139 enrollments for 13,425 patients onto 1,534 clinical trials<br><br>- SOC treatment dataset: 86,042 treatment plans for 50,799 patients | Cancer, including breast, lung, lymphoma and leukemia among others | - Unstructured and structured text | No |
| Chowdhury et al., 2024[29] | Mayo Clinic's United Data Platform | Real patient EHR data<br>Structured EHR: Six types of clinical events (diagnosis, medication, allergy, family history of medical condition, lab tests, and admission (e.g., reason for visit)<br>Unstructured EHR: radiology reports<br>Demographics: age and gender. | 180 patients | 50 out of 180 patients were eligible for the clinical trial | Unstructured and structured text | No |
| Devi et al., 2024 | Synthetic | case reports/patient descriptions | 100 patients | Non-small lung cancer | Patient descriptions | Unknown |
| Ferber et al., 2024[32,33] | Synthetic; based on fictional patient vignettes | Synthetic oncology-focused patient EHRs (describe patient diagnoses, comorbidities, molecular data, imaging descriptions from staging CT or MRI scans, patient history) | 51 records | Cancer, including lung adenocarcinoma | Unstructured text | Yes |
| Gueguen et al, 2025[34] | Local Molecular Tumor Board at Centre Léon Bérard, France | Real tumor board information on sequential patients; clinical data from full EHR and molecular data from molecular programmes | 34 patients in the LLM subset | Mixed adult solid tumors | Unstructured text | Yes; anonymized data available at: https://github.com/crcl-tm2/trialmatch-tool-evaluation/blob/master/artifacts/data_raw/formatted_data.csv |
| Gui et al, 2025[35] | The Hepatology Department of the First Affiliated Hospital of Guangxi University of Chinese Medicine | Real, de-identified narrative admission notes | 16,000 notes from over the last 10 years | Hepatopathy | Unstructured text | Yes; can be obtained from corresponding author upon reasonable request |
| Gupta et al., 2024[14] | Institutional clinical research data warehouse from a single cancer center | Real, de-identified EHR and clinical trial enrollment information, including: assessment & plan note, brief op note, consults, discharge instructions/ summary, h&p, op note, OR surgeon, procedures, progress notes, rad onc simulation, rad onc weekly review | - Q&A Data: Notes from 50 patients<br>- Clinical Data: Notes from over 5790 patients.<br>- To evaluate patient-trial matching: 98 cancer patients<br>- To evaluate trial-patient matching: ~ 1-3 patients who enrolled in each trial, 5-21 who did not | Cancer, including breast and lung | Unstructured text | No |
| Lai et al., 2024[39] | The Pancreas Center | Real de-identified medical oncology note, and surgical oncology note if available, from each patient who was screened for clinical trials at the Pancreas Center between January and May 2024 | 32 patients | Pancreatic cancer; 19 out of 24 patients in the test set were eligible for at least one trial | Unstructured text | Not Reported |
| SIGIR 2016[20] | Adopted from TREC CDS[55] | Real case reports | 60 case reports | Diagnosis varied | Unstructured text | Yes<br>- Available at https://data.csiro.a |
| | | | | | | u/collection/csiro:17152 |
| TREC 2021CT,[17] TREC 2022 CT[18] | "cases created by individuals with medical training" | Synthetic admissions notes | - TREC 2021 CT: 75 topics<br>- TREC 2022 CT: 50 topics | Diagnosis varied | Unstructured text | Yes<br>- TREC 2021 CT: Available at http://www.trec-cds.org/2021.html<br>- TREC 2022 CT: Available at http://www.trec-cds.org/2022.html |
| TREC 2023 CT[19] | Synthetic | Synthetic questionnaire data | 451,538 documents for 50 patients | Glaucoma, anxiety, COPD, breast cancer, COVID-19, rheumatoid arthritis, sickle cell anemia, type 2 diabetes | Structured text | Yes<br>Available at https://www.trec-cds.org/2023.html |
| Unlu et al., 2024[46] | The ongoing Co-Operative Program for Implementation of Optimal Therapy in Heart Failure Trial | Real EHR data over 2 years from ongoing trial including: progress notes, discharge summaries, history and physical, telephone encounters, notes to patients sent through the portal | 1891 patients | High rates of symptomatic heart failure | Unstructured text | No |
| Wong, 2023[47] | EHR from a collaborating health system | NLP-based extraction on structured data from EHR | 523 patient-trial enrollment pairs | Patient-trial enrollment pairs pulled from historical enrollment data | Structured text | No |
| Woo, 2024 (MIMIC-III based)[48] | LLM-generated based on MIMIC-III[56] | Llama-3.1-70B-Instruct was prompted to generate eligibility-criteria-based Q&A pairs based on discharge summaries from MIMIC-III | 1,000 questions | Diagnosis varied | Structured text | Yes<br>https://physionet.org/content/mimic-ext-synth-trial-question/1.0.0/ |
| Woo, 2024 (MIMIC-IV based)[48] | Human-generated based on MIMIC-IV[57] | **Researchers wrote 23 questions similar to eligibility criteria modeled after the ARISTOTLE Apixaban vs. Warfarin trial, then manually annotated answers from each MIMIC-IV discharge summary** | 23 questions resembling eligibility criteria with a random sample of 100 patient notes from MIMIC-IV | Diagnosis varied | Structured text | Yes<br>https://physionet.org/content/mimic-iv-ext-apixaban-trial/1.0.0/ |
| Yuan et al., 2024[49] | Stroke patient database | Real, longitudinal prescription and medical claims data (diagnoses, procedures, medications) | Claims data for 825 patients | Each patient enrolled in 1/6 stroke trials | Unstructured text | Not Reported |
| Zhuang et al., 2024[50] | ChatGPT<br>TREC 2022 CT<br>TREC 2023 CT | - ChatGPT data: Patient descriptions generated for randomly selected clinical trials | ~20,000 synthetic patient-trial pairs (25,000 total once they combined it with 5,000 patient description-trial pairs from TREC 2022 and 2023) | Diagnosis varied | Unstructured Text | Not Reported |
| Zihang et al., 2025[51] | LLM-generated | Simulated answers to eligibility questionnaires | 85 patients | Diagnosis varied | Structured | Available upon request |

*2018 n2c2 denotes 2018 National Natural Language Professing Clinical Challenges cohort; COPILOT-HF, Cooperative Program for ImpLementation of Optimal Therapy in Heart Failure; MGB, Mass General Brigham; MSKCC, Memorial Sloan Kettering Cancer Center; PAH, pulmonary arterial hypertension; RAG, retrieval-augmented generation; SIGIR 2016, Special Interest Group on Information Retrieval 2016 cohort; TCGA, The Cancer Genome Atlas; TREC 2021/2022/2023 CT, Text REtrieval Conference Clinical Trials Track

Source: Data compiled from the studies included in this systematic review.

**Table 3. Clinical Trial Data Used for Training and Testing of LLMs**

| Study Dataset | Data Source | Data Description | Diagnoses Covered in Data Set |
|---|---|---|---|
| Beattie et al., 2024[27] | Phase II trial investigating hypofractionated radiation therapy for head and neck cancer | 14 criteria identified based on trial inclusion/exclusion criteria for evaluation using the LLMs | Head and neck cancer |
| Cerami et al., 2024[28] | Dana-Farber Harvard Cancer Center clinical trial database; ClinicalTrials.gov | Therapeutic clinical trials open at Dana-Farber Harvard Cancer Center from January 2012 to June 2024; 500 trials open for cancer diagnoses on October 22, 2024 | Cancer |
| Chowdhury et al., 2024[29] | ClinicalTrials.gov | Five clinical trials (NCT02008357, NCT04468659, NCT02669433, NCT01767909, NCT02565511) | Cognitive disorders, such as Alzheimer's disease and Dementia with Lewey Bodies |
| Devi et al., 2024[32] | Clinicaltrials.gov | 10 U.S. drug-only interventional clinical trials | Non Small Cell Lung Cancer |
| Ferber et al., 2024[33] | ClinicalTrials.gov | 105,600 real clinical trials filtered for cancer collected on May 13, 2024 | Cancer |
| Gui et al, 2025[35] | Six real-world clinical trials on hepatopathy: ChiCTR2100044187, NCT04353193, NCT04850534, NCT03911037, NCT01311167, NCT04021056 | "Selected 58 criteria assessable using information typically documented in admission notes" | chronic liver failure, cirrhosis, hepatocellular carcinoma |
| Gupta et al., 2024[14] | ClinicalTrials.gov | For patient-trial matching: set of 10 trials for each patient, each for the same cancer type that were actively recruiting when patient enrolled in clinical trial (1/10 of those trials was one in which the patient actually enrolled)<br><br>For trial-patient matching: Set of 36 clinical trials that all recruited patients from the same institution | Cancer |
| Lai et al., 2024[39] | ClinicalTrials.gov | Real trial data from nine ongoing clinical trials at the Pancreas Center | Pancreatic cancer |
| Lin et al., 2024[40] | - ClinicalTrials.gov<br>- PubMed Central<br>- TrialAlign | ClinicalTrials.gov: 467,944 trials, real data<br>- ChiCTR (China): 76,186 trials, real data<br>- TrialAlign: 14 sources of trial documents | Diagnoses varied |
| PROTECTOR1 dataset[58] | ClinicalTrials.gov | 764 phase 3 cancer trials | Cancer |
| Ruan et al., 2024[44] | ClinicalTrials.gov | 301 trials | Fatty liver disease |
| SIGIR 2016[20] | ClinicalTrials.gov | 204,855 trials | Diagnosis varied |
| TREC 2021CT,[17]<br>TREC 2022 CT[18] | ClinicalTrials.gov | 375,581 clinical trial descriptions | Diagnosis varied |
| TrialAlign[40] | 14 sources, including ClinicalTrials.gov and ChiCTR (China) | 793,279 trial documents<br>1,113,207 scientific papers related to clinical trials | Diagnosis varied |
| Unlu et al., 2024[46] | Data from the ongoing Co-Operative Program for Implementation of Optimal Therapy in Heart Failure Trial in the Microsoft Dynamics 365 used by staff for patient screening | 13 trial criteria | Heart failure |
| Wong, 2023[47] | ClinicalTrials.gov | 53 treatment-oriented, interventional trials | Cancer |
| Yuan et al., 2024[49] | ClinicalTrials.gov | 6 trials | Stroke |
| Zihang et al., 2025[51] | ClinicalTrials.gov | 579 clinical trial registration entries with Fudan University as the sponsor; 562 pre-recruitment questionnaires were generated based on 562 trials with complete data | Diagnosis varied |

*MSKCC denotes Memorial Sloan Kettering Cancer Center. All articles that used TREC CT patient data sets pulled trial protocol data from clinicaltrials.gov unless otherwise specified.

Source: Data compiled from the studies included in this systematic review.

**Figures**

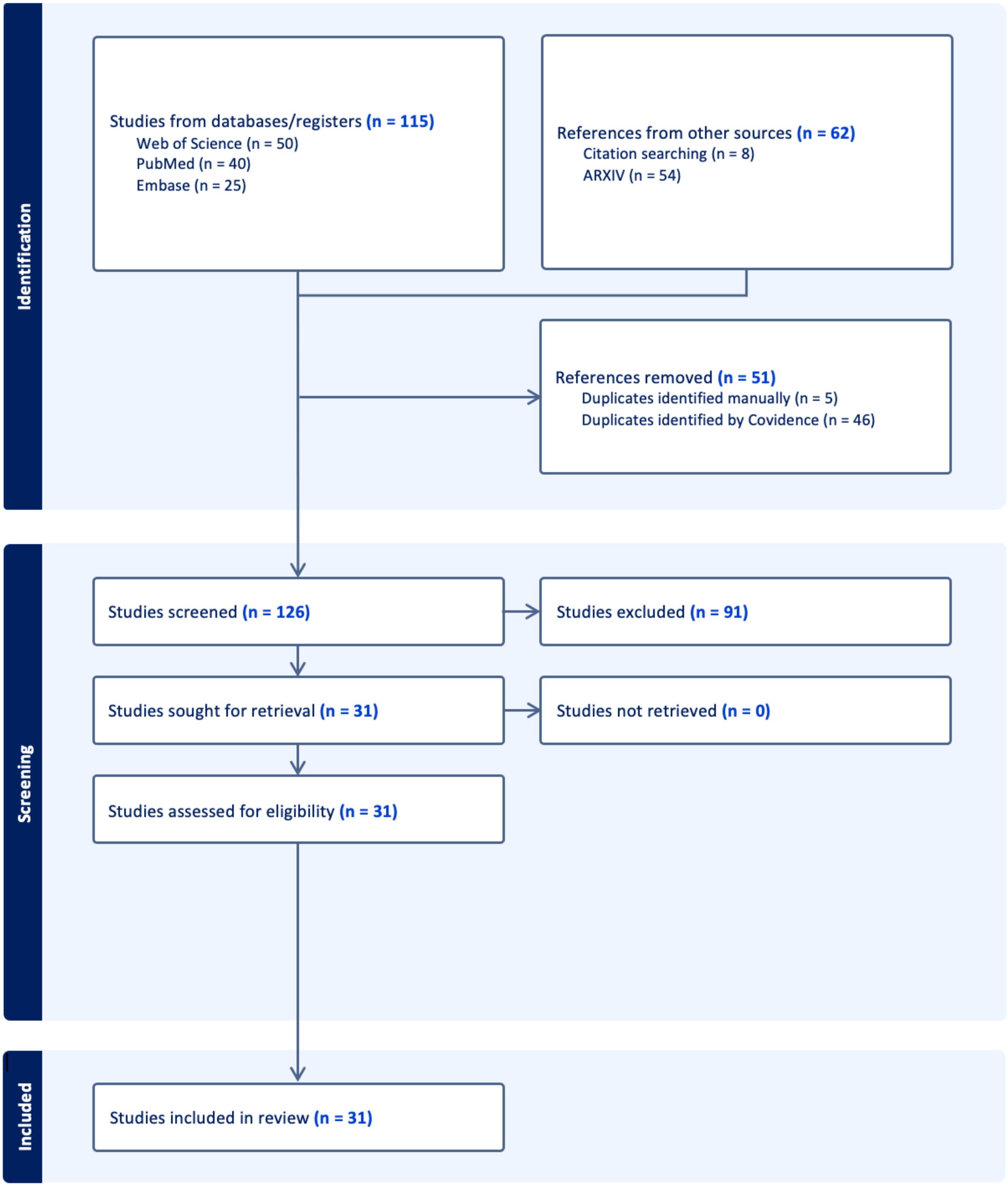

*Figure 1. Flowchart depicting the systematic review search strategy and study selection according to PRISMA guidelines.* Searches across three academic databases and one preprint server yielded 126 unique journal articles. Records were screened for eligibility, with exclusions primarily due to duplication or failure to meet predefined inclusion criteria. Source: Covidence systematic review software, Veritas Health Innovation, Melbourne, Australia. Available at www.covidence.org.

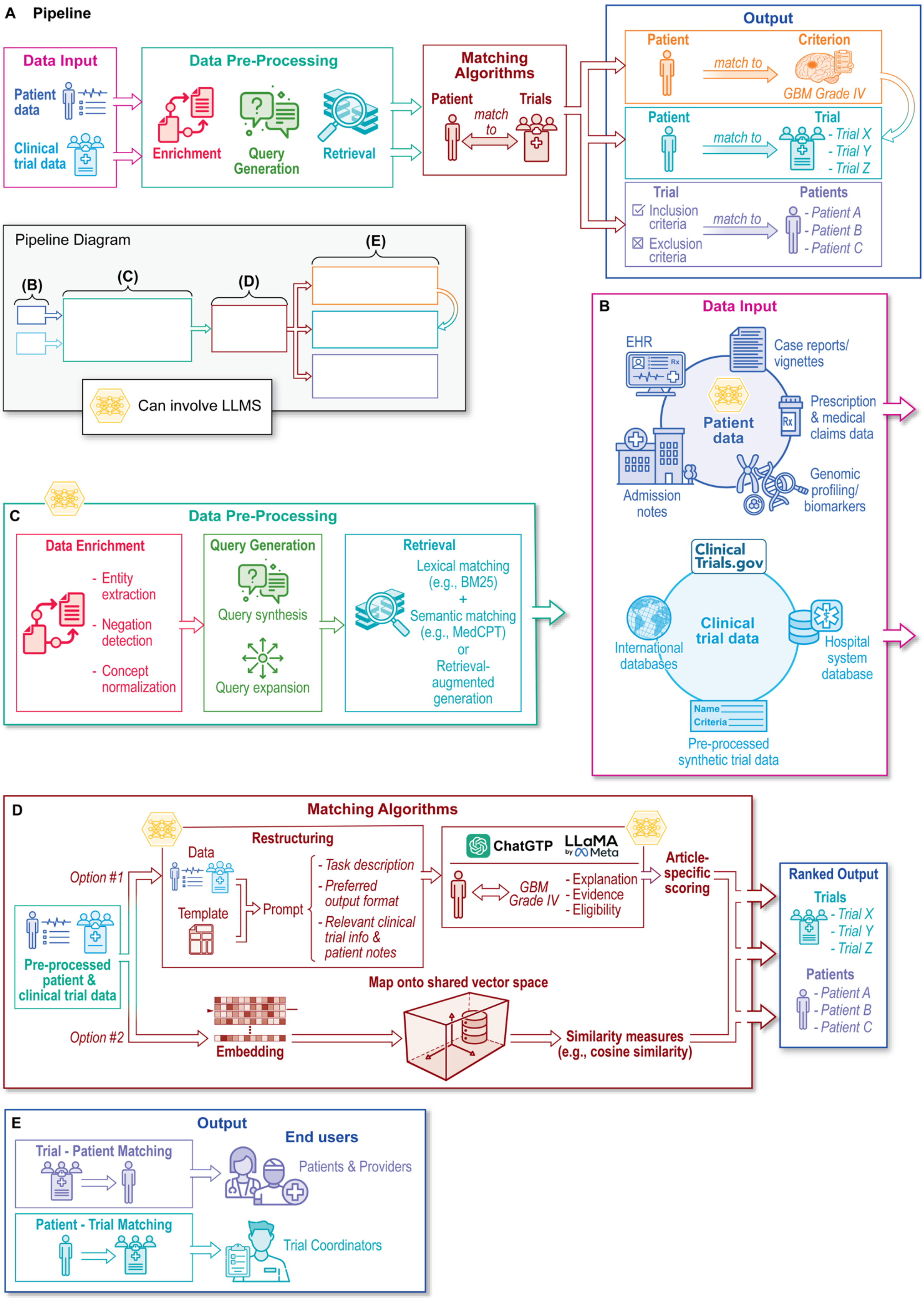

A Pipeline
Data Input
Patient data
Clinical trial data
Data Pre-Processing
Enrichment
Query Generation
Retrieval
Matching Algorithms
Patient
Trials
match to
Output
Patient
Criterion
match to
GBM Grade IV
Patient
Trial
match to
- Trial X
- Trial Y
- Trial Z
Trial
Patients
Inclusion criteria
Exclusion criteria
match to
- Patient A
- Patient B
- Patient C
Pipeline Diagram
(B)
(C)
(D)
(E)
Can involve LLMS
B
Data Input
EHR
Case reports/ vignettes
Patient data
Prescription & medical claims data
Admission notes
Genomic profiling/ biomarkers
Clinical Trials.gov
International databases
Clinical trial data
Hospital system database
Name
Criteria
Pre-processed synthetic trial data
C
Data Pre-Processing
Data Enrichment
- Entity extraction
- Negation detection
- Concept normalization
Query Generation
Query synthesis
Query expansion
Retrieval
Lexical matching (e.g., BM25)
+
Semantic matching (e.g., MedCPT)
or
Retrieval-augmented generation
D
Matching Algorithms
Option #1
Restructuring
Data
Template
Prompt
- Task description
- Preferred output format
- Relevant clinical trial info & patient notes
ChatGTP
LLaMA by Meta
GBM Grade IV
- Explanation
- Evidence
- Eligibility
Article-specific scoring
Pre-processed patient & clinical trial data
Map onto shared vector space
Option #2
Embedding
Similarity measures (e.g., cosine similarity)
Ranked Output
Trials
- Trial X
- Trial Y
- Trial Z
Patients
- Patient A
- Patient B
- Patient C
E
Output
End users
Trial - Patient Matching
Patients & Providers
Patient - Trial Matching
Trial Coordinators

*Figure 2. Schematic Overview of Common Trial Matching Pipeline Designs.* (A) Key components of the reviewed trial matching pipelines include data input, data pre-processing, matching algorithms, and the output, which takes the form of patient-criterion, patient-trial, and/or trial-patient matches. (B) Each article used its own combination of patient and trial data. (C) Various data enrichment methods are employed on trial or patient data to improve the generation of queries, which are then used to retrieve the patient information most relevant to a given criterion or trial. In most pipelines, this patient and trial information is then restructured and inserted into a prompt to be fed into an LLM. (D) The first and most common matching pipeline involves feeding prompts with patient and trial data to an LLM, the output of which is inputted into article-specific scoring algorithms to generate ranked outputs. The second pipeline involves embedding specific sets of patient and trial data (e.g. patient summaries and eligibility criteria) into a shared vector space, then using similarity measures like cosine similarity to rank patients or trials. (E) While patient-trial matching can be used by patients and providers to identify trials for potential enrollment, trial-patient matching would be used by trial coordinators to identify patients to enroll in a given trial.

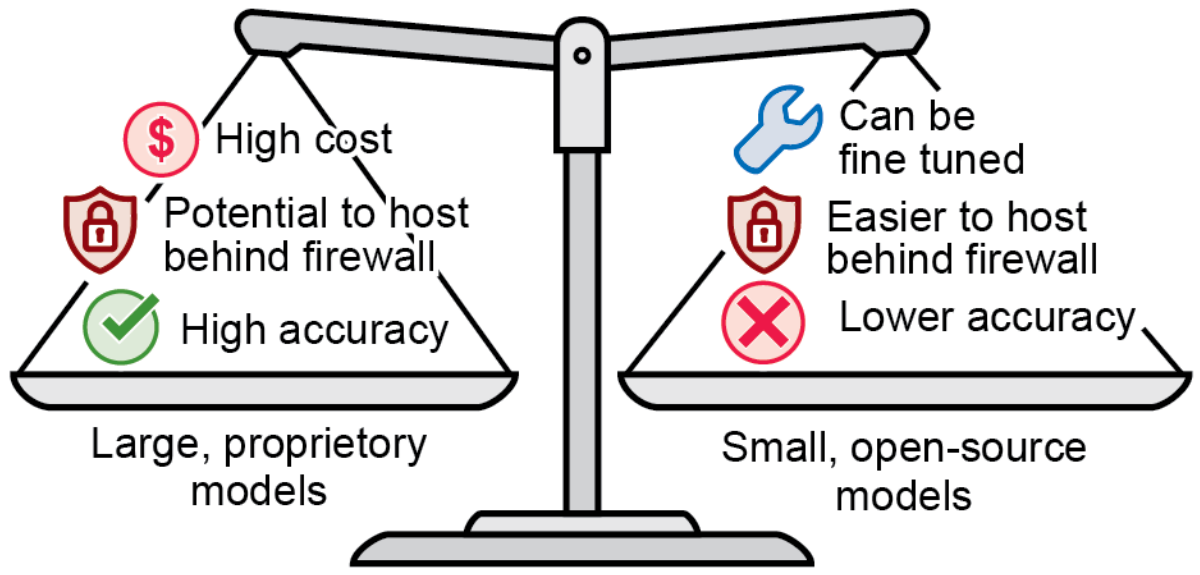

*Figure 3. Consideration of Factors Influencing Choice of Proprietary or Open-Source Models*. Key considerations for choosing between larger, proprietary models and smaller, open-source models include performance, cost and ability to host behind hospital firewalls for safety of protected health information.